\documentclass[conference]{IEEEtran}
\IEEEoverridecommandlockouts

\usepackage{cite}
\usepackage{amsmath,amssymb,amsfonts}
\usepackage{array}
\usepackage{algorithmic}
\usepackage{multirow}
\usepackage{multicol}
\usepackage{textcomp}
\usepackage{xcolor}
\usepackage{subfig}
\usepackage{graphicx, caption, subcaption}
\usepackage{adjustbox}
\usepackage{makecell}
\def\BibTeX{{\rm B\kern-.05em{\sc i\kern-.025em b}\kern-.08em
    T\kern-.1667em\lower.7ex\hbox{E}\kern-.125emX}}
\begin{document}

\title{Segmentation-Driven Initialization for Sparse-view 3D Gaussian Splatting}

\author{Yi-Hsin~Li, Thomas Sikora,~\IEEEmembership{Senior Member,~IEEE}, Sebastian~Knorr,~\IEEEmembership{Senior Member,~IEEE,} \\Mårten~Sjöström,~\IEEEmembership{Senior Member,~IEEE,}\vspace{-3ex}
\thanks{Manuscript received June 2025. This project has received funding from the European Union’s Horizon 2020 research and innovation program under the Marie Skłodowska-Curie grant agreement No 956770, and by Mid Sweden University internal funding. Computations were enabled by NAISS, partly funded by Swedish Research Council (2022-06725), and by High Performance Computing Center North (HPC2N) at Umeå University. \textit{(Corresponding authors: Mårten~Sjöström.)}}
\thanks{Yi-Hsin~Li is with Department of Computer and Electrical Engineering, Mid Sweden University, Sundsvall, 85170, Sweden, and also with Department of Telecommunication Systems, Technical University of Berlin, Berlin, 10587, Germany (e-mail: yi-hsin.li@miun.se).}
\thanks{Sebastian~Knorr is with School of Computing, Communication and Business, Hochschule für Technik und Wirtschaft Berlin, Berlin, 12459, Germany (e-mail: sebastian.knorr@htw-berlin.de).}
\thanks{Mårten~Sjöström is with Department of Computer and Electrical Engineering, Mid Sweden University, Sundsvall, 85170, Sweden (e-mail: Marten.Sjostrom@miun.se).}
\thanks{Thomas~Sikora is with Department of Telecommunication Systems, Technical University of Berlin, Berlin, 10587, Germany (e-mail: thomas.sikora@tu-berlin.de).}
}

\maketitle

\begin{abstract}
Sparse-view synthesis remains a challenging problem due to the difficulty of recovering accurate geometry and appearance from limited observations. While recent advances in 3D Gaussian Splatting (3DGS) have enabled real-time rendering with competitive quality, existing pipelines often rely on Structure-from-Motion (SfM) for camera pose estimation—an approach that struggles in genuinely sparse-view settings. Moreover, several SfM-free methods replace SfM with multi-view stereo (MVS) models, but generate massive numbers of 3D Gaussians by back-projecting every pixel into 3D space, leading to high memory costs.
%
We propose Segmentation-Driven Initialization for Gaussian Splatting (SDI-GS), a method that mitigates inefficiency by leveraging region-based segmentation to identify and retain only structurally significant regions. This enables selective downsampling of the dense point cloud, preserving scene fidelity while substantially reducing Gaussian count. Experiments across diverse benchmarks show that SDI-GS reduces Gaussian count by up to 50\% and achieves comparable or superior rendering quality in PSNR and SSIM, with only marginal degradation in LPIPS. It further enables faster training and lower memory footprint, advancing the practicality of 3DGS for constrained-view scenarios.

\end{abstract}


\section{Introduction}
Sparse-view rendering, the task of synthesizing novel views from a limited number of input images, has become a critical challenge in 3D computer vision. It is especially relevant in real-world applications such as robotics \cite{zhu20243dgsrobotics}, augmented reality \cite{jiang2024vrgs}, and medical imaging \cite{zhu2024endogs, zhou2023tiavox}, where acquiring densely sampled views is either impractical or impossible. 
This paper investigates how segmentation can be leveraged within Gaussian Splatting to intelligently select a compact yet sufficient set of Gaussians for efficient reconstruction from sparse views.

A wave of progress in view synthesis was sparked by Neural Radiance Fields (NeRF) \cite{nerf}. It models scenes as continuous volumetric fields via implicit neural networks. While initially reliant on dense inputs, several variants \cite{chibane2021SRF, truong2023SPARF, Niemeyer2021Regnerf, meuleman2023progressivenerf} have sought to extend NeRF’s capabilities to sparse-view settings. These methods introduce auxiliary constraints, such as geometry priors, stereo-inspired similarity, or multi-view correspondence losses, to guide reconstruction from limited views. However, NeRF and its variants are notoriously slow to train and render, making them unsuitable for real-time or large-scale scenarios.

Recently, view synthesis and radiance field modelling has emerged with a powerful alternative in 3D Gaussian Splatting (3DGS) \cite{kerbl_3d_2023}. It represents a scene using a set of spatially distributed 3D Gaussians, each with learnable attributes such as position, opacity, and color. This explicit, point-based representation enables high-quality rendering in real time and has rapidly become a new standard for efficient view synthesis.

3DGS exposes a critical limitation when applied to sparse-view scenarios. Typically, it depends on Structure-from-Motion (SfM) \cite{westoby2012sfm} to estimate camera poses and initialize 3D point distributions. SfM, which relies on robust feature matching across multiple views, becomes fragile when inputs are sparse, often resulting in erroneous geometry and pose estimates. Several recent pipelines claim to operate under sparse-view conditions, but they assume known camera poses as input. In practice, these poses are not obtained from the sparse views themselves, but are instead computed on densely sampled video sequences using Structure-from-Motion (e.g., COLMAP \cite{schoenberger2016sfm, schoenberger2016mvs}). Although only a few camera views are selected for point cloud generation, the underlying poses originate from the full set of dense views. This reliance on precomputed dense-view poses undermines the true sparsity assumption and raises concerns about the fairness and generality of such evaluations.

Recent methods \cite{Fu_2024_CVPR, fan2024instantsplat} have proposed SfM-free pipelines to overcome the limitations of SfM in sparse-view settings. They estimate camera poses directly from the input views using multi-view stereo networks. While these approaches eliminate the dependency on SfM, they introduce a new inefficiency: dense, pixel-wise lifting of every image into 3D, regardless of structural relevance. This uniform initialization inflates the number of Gaussians, especially leading to redundant representations in flat or low-texture regions. The result is excessive memory usage and slower rendering, which undermines the efficiency gains expected from SfM-free pipelines. 

We observe that improved initialization plays a critical role in mitigating the drawbacks of dense, unstructured initial point clouds. Our previous work on 2D Gaussian-based regression \cite{li_segmentation-based_2023, adptsmoe} demonstrated that region-based segmentation can effectively reduce redundancy while preserving important structures. Motivated by this, we extend the segmentation-driven paradigm from 2D to the 3D Gaussian Splatting domain. Our method leverages 2D region-based cues to identify consistent, structurally meaningful areas across views and guide selective downsampling. This pre-filtering step yields compact, geometry-aware initialization, retaining essential structure while reducing unnecessary overhead. As a result, our method improves memory and runtime efficiency without compromising visual fidelity. Our contributions are as follows:

\begin{itemize}
    \item We introduce a segmentation-driven strategy for initializing 3D Gaussians, reducing computational and memory demands in sparse-view rendering.

    \item Our method reduces memory usage by up to 50\% while preserving image quality and maintaining the fast training performance characteristic of SfM-free pipelines—achieving competitive PSNR and SSIM with minimal LPIPS degradation.

    \item We evaluate against SfM-free and SfM-based baselines, demonstrating improved efficiency and competitive quality across diverse sparse-view scenarios.

\end{itemize}

\begin{figure*}[th]
    \centering
    \includegraphics[width=0.99\linewidth,trim={1.8cm 2.7cm 2.0cm 2.8cm},clip]{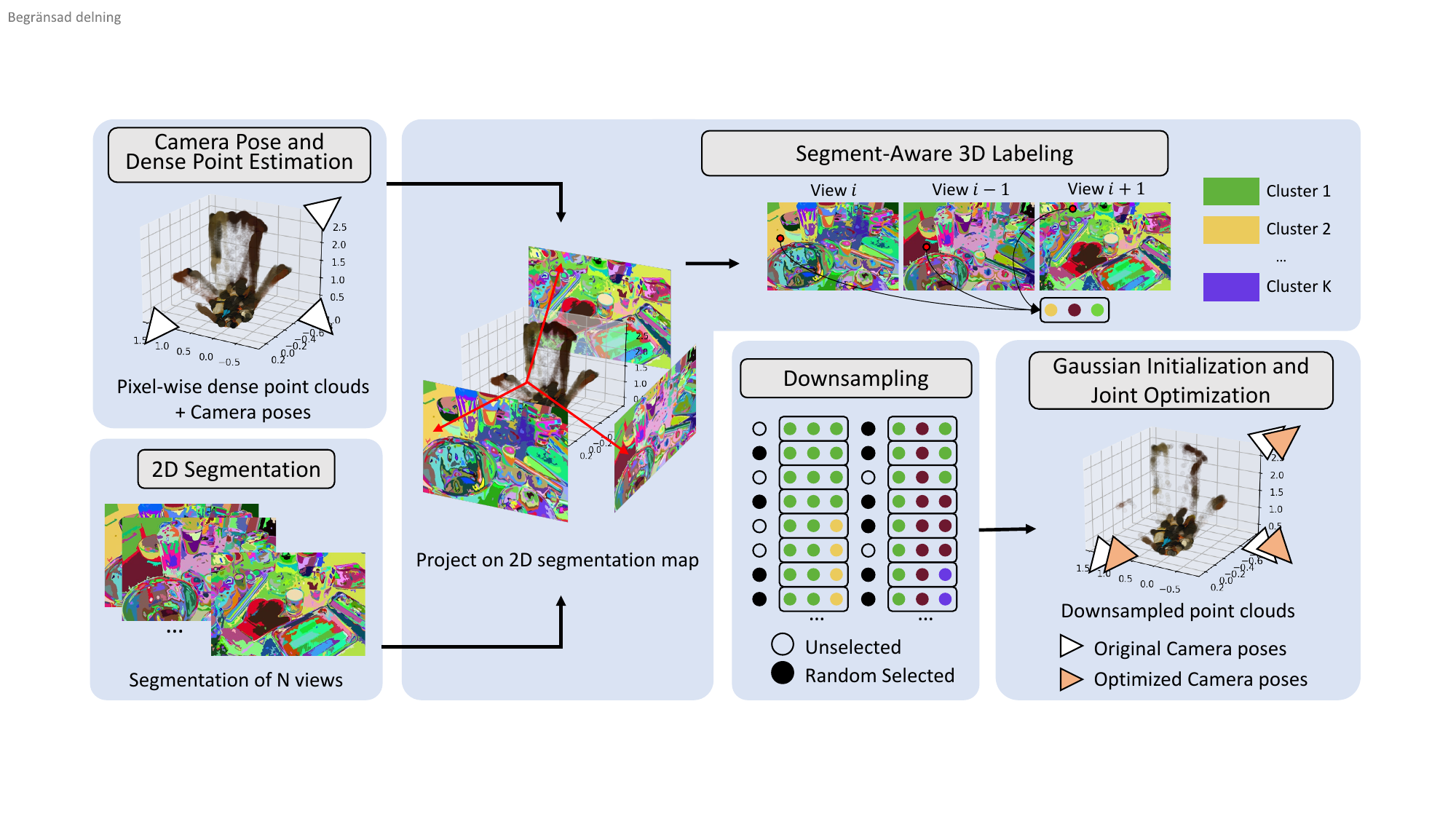}
    \caption{Overview of our segmentation-driven initialization pipeline for sparse-view 3D Gaussian Splatting. Given sparse input views, we estimate camera poses and lift all image pixels into a dense 3D point cloud. We apply region-based segmentation on each image and propagate these segmentations across views to construct segment-aware 3D labels. These labels guide a structured downsampling process that prunes redundant points while preserving geometric structure. The resulting filtered points initialize 3D Gaussians, which are jointly optimized with camera poses to produce the final radiance field.}
    \label{fig:1}
\end{figure*}

\section{Related work}
This section reviews three key domains relevant to our work: 3D Gaussian Splatting for sparse-view rendering, segmentation-guided strategies for learning Gaussian parameters, and learning-based camera pose estimation methods. 
First, we emphasize 3D Gaussian Splatting (3DGS) for its rendering efficiency and explicit point-based structure, which make it particularly effective in sparse-view scenarios where volumetric or implicit approaches face challenges with scalability and transparency. Second, segmentation has become an important tool for structuring Gaussian representations. Unlike prior works that typically use semantic segmentation as a training regularizer, our method leverages region-based segmentation at initialization. This motivates a broader investigation into how segmentation can guide the learning of Gaussian parameters throughout different stages. Finally, we review learning-based camera pose estimation methods because our approach demands alternatives to traditional SfM for accurate pose initialization. We recognize that SfM’s reliance on feature matching falters when input views are sparse or challenging.


\subsection{Sparse-view 3D Gaussian Splatting}
Within the 3DGS framework, pipelines designed for sparse-view rendering can be broadly categorized as SfM-based, Hybrid, and SfM-free methods, depending on whether they rely on Structure-from-Motion for camera pose estimation and scene initialization.

\subsubsection{SfM-based Methods}
Several methods in sparse-view rendering rely heavily on SfM, most commonly using COLMAP for estimating camera poses and reconstructing initial point clouds. Few-shot View Synthesis using Gaussian Splatting (FSGS) \cite{zhu2023FSGS} uses COLMAP-derived points and poses to initialize Gaussians, then improves scene coverage through proximity-guided unpooling, inserting new Gaussians between visually distinct ones. Similarly, SparseGS \cite{xiong2023sparsegs} leverages COLMAP outputs and refines novel views through depth correlation and floater pruning guided by rendered depth. CoR-GS \cite{zhang2024corgs} trains two parallel Gaussian radiance fields using SfM-based initialization. It enforces consistency by pruning unmatched Gaussians and using co-rendered pseudo views as mutual supervision during training.

The problem with SfM-based methods is that the initialization becomes unreliable in sparse-view settings due to limited feature correspondences. A common workaround is to estimate poses from densely sampled sequences and then select a sparse subset for training. This practice raises concerns about the fairness and validity of such evaluations. These limitations motivate the development of SfM-free alternatives for both pose estimation and Gaussian initialization.

\subsubsection{Hybrid Methods}
Hybrid approaches retain SfM for camera pose estimation but eliminate its role in point cloud initialization. DNGaussian \cite{li2024dngaussian} replaces SfM-derived geometry with randomly initialized Gaussians and introduces depth-based regularization using locally and globally normalized rendered depth maps. However, it continues to rely on COLMAP for pose estimation. RegSegField \cite{kai2024regsegfield} relies on SfM-derived camera poses but initializes from randomly distributed 3D Gaussians rather than SfM-generated geometry. It incorporates semantic segmentation during training to guide hierarchical refinement, effectively bypassing traditional SfM-based point cloud initialization. While hybrid methods attempt to reduce reliance on SfM, they still depend on it for critical components like camera pose estimation. This residual dependence limits their applicability in scenarios where SfM is unreliable or infeasible. Our work explicitly targets full removal of SfM, which motivates us to compare against methods that either fully adopt SfM or entirely avoid it, rather than hybrids that inherit its limitations.

\subsubsection{SfM-free Methods}
To overcome the fragility and inefficiency of SfM, several recent methods have explored SfM-free alternatives. COLMAP-free GS \cite{Fu_2024_CVPR} lifts 2D pixels into 3D using monocular depth predictions and camera intrinsics, forming local Gaussians for each view. These Gaussians are aligned across views via affine transformations optimized with a rendering loss. To improve coverage, the method progressively densifies regions with high reconstruction error. However, this approach struggles with large camera motions, where affine alignment becomes unreliable and reconstruction quality degrades.

InstantSplat \cite{fan2024instantsplat} further demonstrates the viability of SfM-free pipelines by jointly estimating camera poses and initializing Gaussians from dense pixel-wise point clouds. Although it incorporates confidence-aware filtering to remove redundant points in co-visible regions, the resulting point cloud remains overly dense, particularly in non-overlapping and low-texture areas. This leads to significant memory overhead and limits scalability in truly sparse-view conditions.

Rather than relying solely on confidence in co-visible areas, our method applies region-based segmentation to guide the filtering process. We identify structurally meaningful regions across views and selectively retain representative points. This segmentation-driven filtering reduces redundancy at the source, yielding more compact and spatially coherent Gaussian distributions.

\subsection{Segmentation-Guided Methods}
Segmentation has been incorporated into the family of Gaussian-based models for both training-time supervision and initialization.

\subsubsection{Training-Based Segmentation}
Several methods incorporate segmentation during training to regularize Gaussian supervision. SAM3D \cite{zhang2023sam3d} lifts 2D masks from Segment Anything (SAM) into 3D and merges them via mesh-based alignment and bidirectional refinement. Gaussian Grouping \cite{ye2024gaussian_grouping} associates SAM masks across views using identity tracking, then constrains the rendered Gaussians to match these labels. RegSegField \cite{kai2024regsegfield}, as discussed earlier, learns view-invariant segment descriptors from 2D SAM masks to guide hierarchical refinement during optimization.

Previous training-based segmentation approaches operate under dense or densifying conditions and depend on high-level semantic labels. In contrast, our method targets sparse-view settings where no densification is applied, using lightweight, region-based segmentation to inform Gaussian initialization rather than supervision.


\subsubsection{Initialization-Based Segmentation}
Certain prior efforts use segmentation for guiding Gaussian initialization.
 S-SMoE \cite{li_segmentation-based_2023} and AS-SMoE \cite{adptsmoe} leverage region-based segmentation to reduce kernel redundancy in Steered Mixture of Experts (SMoE), an edge-aware Gaussian-based regression model. These approaches focus on preserving high-frequency structures while simplifying homogeneous regions.

Although limited to 2D, these works demonstrate that region-aware initialization offers meaningful trade-offs between compactness and fidelity. Recognizing the conceptual alignment between SMoE and 3D Gaussian Splatting, we extend this segmentation-driven strategy to the 3D domain. Our method leverages cross-view region consistency to guide 3D Gaussian initialization, enabling efficient, structure-aware modeling from the outset without relying on semantic labels or progressive densification.

\subsection{Learning-Based Camera Pose Estimation Methods}

Structure-from-Motion (SfM) extracts camera poses and sparse 3D points through feature matching and incremental optimization. While reliable, SfM depends on robust feature detection and careful calibration, limiting flexibility and speed.

Some works focus on improving SfM’s components rather than bypassing it entirely. For instance, SuperGlue \cite{sarlin20superglue} enhances feature matching via learning-based attention, boosting robustness—but it remains tethered to SfM pipelines.

Breaking away, PoseNet \cite{kendall2015posenet} introduced a bold alternative: directly regressing camera poses from single images using convolutional neural networks, eliminating reliance on feature matching or multi-view geometry. Building on this, MapNet \cite{mapnet2018} incorporates geometric constraints and temporal context, refining pose estimation accuracy without SfM.

Despite progress, many learning-based pose methods focus solely on camera poses, often overlooking scene geometry or producing sparse outputs. DUSt3R~\cite{dust3r_cvpr24} breaks this mold by combining pose estimation and dense 3D reconstruction in a Transformer-based, end-to-end pipeline. MASt3R~\cite{mast3r}, built on DUSt3R’s backbone, enhances both accuracy and efficiency by introducing local dense feature proposals and a dedicated matching loss. By jointly regressing dense points and camera poses from uncalibrated images, it delivers a fully learning-driven, SfM-free solution, making it the method of choice for our pipeline.

\begin{figure}[t]
\centering
\vspace{-3mm}
\subfloat[Input RGB image]{%
  \includegraphics[width=0.48\linewidth]{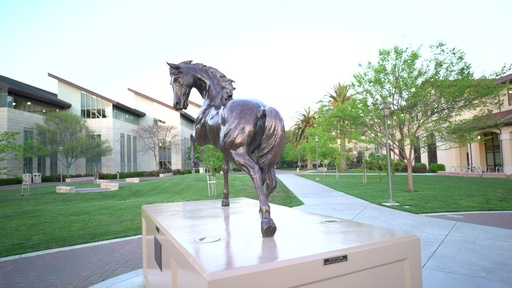}
}
\hfill
\subfloat[Segmentation map]{%
  \includegraphics[width=0.48\linewidth]{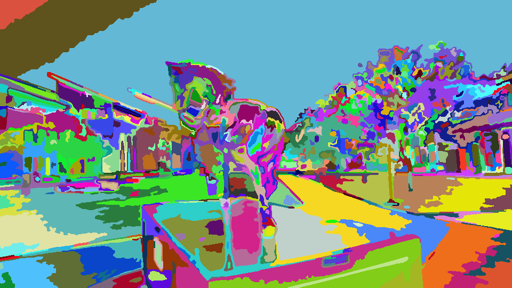}
}
\\
\subfloat[Retained mask]{%
  \includegraphics[width=0.48\linewidth]{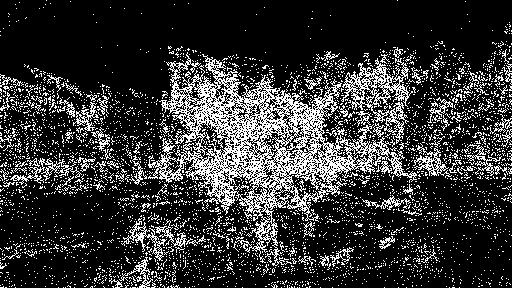}
}
\hfill
\subfloat[Downsampled 3D points]{%
  \includegraphics[width=0.48\linewidth]{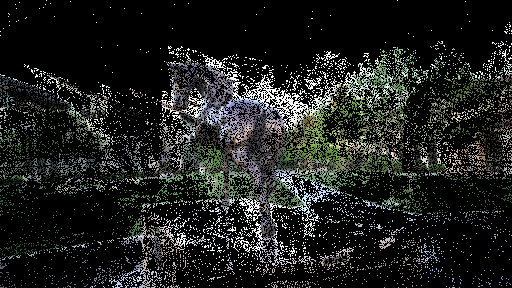}
}

\caption{Visualization of segmentation-guided downsampling. (a) Input RGB image; (b) region-based segmentation map; (c) retained pixel mask after stratified sampling; (d) final downsampled 3D points projected onto the image plane. Redundant points in flat areas (e.g., sky) are removed, while structural details are preserved by retaining more points in high-frequency regions.}
\label{fig:2}
\vspace{-1em}
\end{figure}

\section{Method}

An overview of our SDI-GS pipeline is shown in Fig.~\ref{fig:1}. The process begins with Dense and Unconstrained Stereo 3D Reconstruction \cite{mast3r} for estimating camera poses and generating dense point clouds from the input views. Subsequently, 2D region-based segmentation is performed on each view, and the dense point clouds are projected across views to construct segment-aware 3D labels. These labels guide a structured downsampling process that reduces redundancy while maintaining the scene’s structural fidelity. The downsampled points serve as the basis for initializing 3D Gaussians, which are then refined jointly with camera poses through an optimization procedure. The following subsections describe each component in detail.

\subsection{Camera Pose and Dense Point Estimation}
Given a sparse set of RGB views $\{ I_1, I_2, \dots, I_N \}$, we employ MASt3R to estimate both the relative camera poses $\{(R_i, t_i)\}_{i=1}^{N}$ and dense pixel-wise 3D point clouds. For each pixel $(u, v)$ in view $I_i$, MASt3R lifts it into a 3D point $\mathbf{x}_i(u,v) \in \mathbb{R}^3$, forming a 3D point set $X_i = \{ \mathbf{x}_i(u,v) \}_{(u,v) \in I_i}$ per view. The full unfiltered point set $\mathcal{P}$ across all views is:
\begin{equation}
\mathcal{P} = \bigcup_{i=1}^{N}  X_i .
\end{equation}

While comprehensive, these lifted point clouds contain redundancy due to overlapping views and homogeneous regions. Hence, downsampling is critical for efficiency and quality.

\subsection{2D Segmentation}

To identify structurally meaningful regions in each view, we adopt a modified DBSCAN algorithm (MDBSCAN) \cite{li_segmentation-based_2023}, which clusters pixels based on color similarity in the RGB space. Given the input image $I_i$, MDBSCAN generates a segmentation map, where each pixel $(u, v)$ in view $I_i$ is assigned a region label:

\begin{equation}
\ell^{2D}_i(u,v) \in \mathbb{N}.
\end{equation}

This region-based segmentation approach offers several advantages over semantic segmentation methods. First, it preserves fine-grained structural details by grouping pixels based solely on local color similarity, without imposing semantic constraints that may oversimplify textures or repetitive patterns. This is particularly important in our setting, where we do not employ subsequent densification or pseudo-view regularization. Therefore, we require high-frequency details to be preserved from the initialization stage.

Moreover, unlike many region-growing methods \cite{shen2016dbscan, zhang_fast_2023, yang_superpixel_2020, neubert_compact_2014, li_superpixel_2015}, MDBSCAN imposes no explicit constraint on segment size. This flexibility is crucial: flat regions can be efficiently represented by large Gaussian components, and unnecessarily subdividing them would introduce redundant initializations. Conversely, high-frequency regions are naturally segmented into finer clusters, ensuring that structurally rich areas are well-captured without manual tuning.

In addition to its representational benefits, MDBSCAN is computationally efficient. It runs in approximately 4 ms per view, making it well-suited for our pipeline, where runtime overhead must be minimized. In contrast, deep learning–based segmentation methods such as Segment Anything \cite{Kirillov2023SAM} typically require around 150 ms per view, which is impractical for our intended efficiency. Overall, MDBSCAN strikes a desirable balance between structural fidelity, computational efficiency, and compatibility with sparse-view training objectives.

\subsection{Segment-Aware 3D Labeling}


As defined in Eq. (1), each $X_i$ is a subset of $\mathcal{P}$ containing 3D points lifted from pixels in view $I_i$. 
Each 3D point $\mathbf{x} \in X_i$ is projected onto two adjacent training views, $I_{i-1}$ and $I_{i+1}$, using projection functions $p_{i-1}, p_{i+1} : \mathbb{R}^3 \rightarrow \mathbb{R}^2$, respectively. These functions map the 3D point $\mathbf{x}$ to its corresponding pixel coordinates in the adjacent views, where segmentation labels $\ell^{2D}_{i-1}$ and $\ell^{2D}_{i+1}$ are defined.

For each $\mathbf{x} \in X_i$, the segment-aware 3D label vector is constructed by aggregating segmentation labels from the source and adjacent views as:
\begin{equation}
\ell^{3D}(\mathbf{x}) = \left[ \ell^{2D}_i, \; \ell^{2D}_{i-1}(p_{i-1}(\mathbf{x})), \; \ell^{2D}_{i+1}(p_{i+1}(\mathbf{x})) \right].
\end{equation}

This label vector encodes how a segment from view $ i $ is preserved or fragmented across viewpoints. To ensure that downsampling preserves a meaningful structure without excessive fragmentation, we construct each 3D label vector using only the segment labels from a source view and its two adjacent views. Incorporating more views would increase label dimensionality and risk of oversegmentation, as small inconsistencies across projections can unnecessarily split coherent regions. In practice, three views provide sufficient discriminative power to separate structurally distinct areas while maintaining computational efficiency (see Section 4.2 for ablation results on label dimensionality).


%

We define a structural cluster at the set of all points in the scene that share an identical 3D label vector:

\begin{equation}
C_j = \{ \mathbf{x} \in \mathcal{P} \mid \ell^{3D}(\mathbf{x}) = \ell_j^{3D} \},
\end{equation}
where $ \ell_j^{3D} $ denotes a unique label vector across the entire set $ \mathcal{P} $.

\subsection{Downsampling}
To reduce redundancy while preserving structural diversity, we perform stratified sampling within each cluster $ C_j $. We retain up to $ N_{\text{max}} $ points per cluster, using sampling with replacement. If a cluster contains fewer points than $ N_{\text{max}} $, we sample only as many as available:
\begin{equation}
\hat{C}_j = \text{RandomSample}(C_j, \min(N_{\text{max}}, |C_j|)).
\end{equation}
where $|C_j|$ denotes the cardinality.
The final downsampled point cloud is the union of sampled points across all clusters:
\begin{equation}
\hat{\mathcal{P}} = \bigcup_j \hat{C}_j.
\end{equation}

This sampling strategy assigns a similar number of points to each structurally distinct region, avoiding overrepresentation of large, uninformative areas.
The core idea is to enforce cross-view consistency: if a segment remains coherent when projected across views, it likely corresponds to a structurally reliable region. In contrast, projections that scatter into dissimilar or fragmented segments indicate inconsistency and are treated separately. This strategy enables us to group 3D points based on consistent low-level appearance patterns across views before downsampling, preserving geometric structure while reducing redundant points.

As illustrated in Fig.~\ref{fig:2}, our segmentation-driven strategy effectively prunes redundant points while preserving structural fidelity. Fig.~\ref{fig:2}(a) shows the original input view, and (b) visualizes the region-based segmentation result. Based on these segments, Fig.~\ref{fig:2}(c) indicates the retained mask after sampling representative points within each region. The final retained points, shown in (d), form a significantly reduced yet structurally consistent point cloud.

Regions with low visual complexity—such as sky or flat surfaces—are typically grouped into large segments, from which only a few points are retained. In contrast, high-frequency regions generate more segments and retain more samples, ensuring finer detail. This targeted reduction eliminates redundancy at the source and provides a strong initialization prior for Gaussian optimization.

\subsection{Gaussian Initialization and Joint Optimization}
Each retained 3D point $ \mathbf{x} \in \hat{\mathcal{P}} $ is used to initialize a 3D Gaussian $G = (\mu, \Sigma, c, \alpha)$, where $\mu$ represents the position, $\Sigma$ the covariance, $c$ the color, and $\alpha$ the opacity. The full set of Gaussians is denoted as $\hat{\mathcal{G}} = \{ G \mid \mathbf{x} \in \hat{\mathcal{P}} \}$.

The initialization and optimization processes adhere to the differentiable rendering framework of 3D Gaussian Splatting (3DGS) \cite{kerbl_3d_2023}. Additionally, we employ the joint optimization strategy introduced in InstantSplat \cite{fan2024instantsplat}, wherein both Gaussian parameters and camera poses are refined simultaneously. The optimization is driven by a photometric loss $ L_{\text{photo}} $, which is formulated as:

\begin{equation}
L_{\text{photo}} = \sum_{i=1}^{N} \left| I_i - R(\hat{\mathcal{G}}, T_i) \right|_2^2,    
\end{equation}
where $ I_i $ is the ground truth image for view $ i $, $ T_i $ is the corresponding camera pose, and $ R(\hat{\mathcal{G}}, T_i) $ denotes the rendered image produced by splatting the Gaussians $ \hat{\mathcal{G}} $ under pose $ T_i $.
This joint optimization strategy ensures high-fidelity view synthesis, facilitating accurate reconstruction even with sparsely sampled input views.




\begin{figure*}[ht]
    \centering
    \begin{tabular}{ccc}
    \raisebox{0.5\height}{\rotatebox{90}{Tank and Temples}} & \rotatebox{90}{\hspace{2em}12 views \hspace{3em}3 views}
    \includegraphics[width=0.95\linewidth]{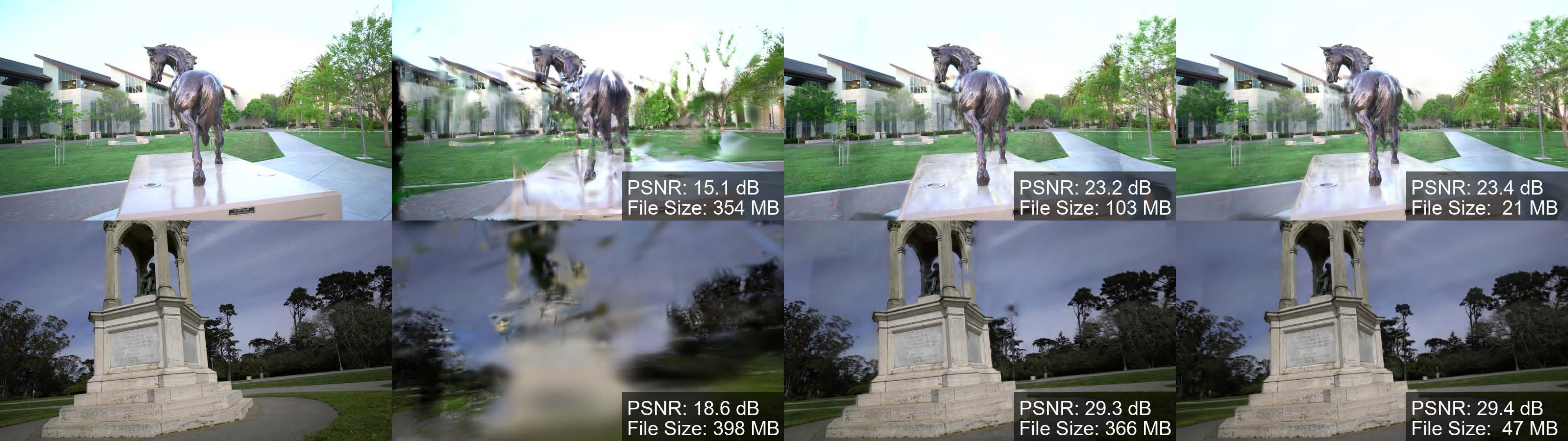} \\
    
    \raisebox{1.0\height}{\rotatebox{90}{MVImgNet}} & \rotatebox{90}{\hspace{2em}12 views \hspace{3em}3 views}
    \includegraphics[width=0.95\linewidth]{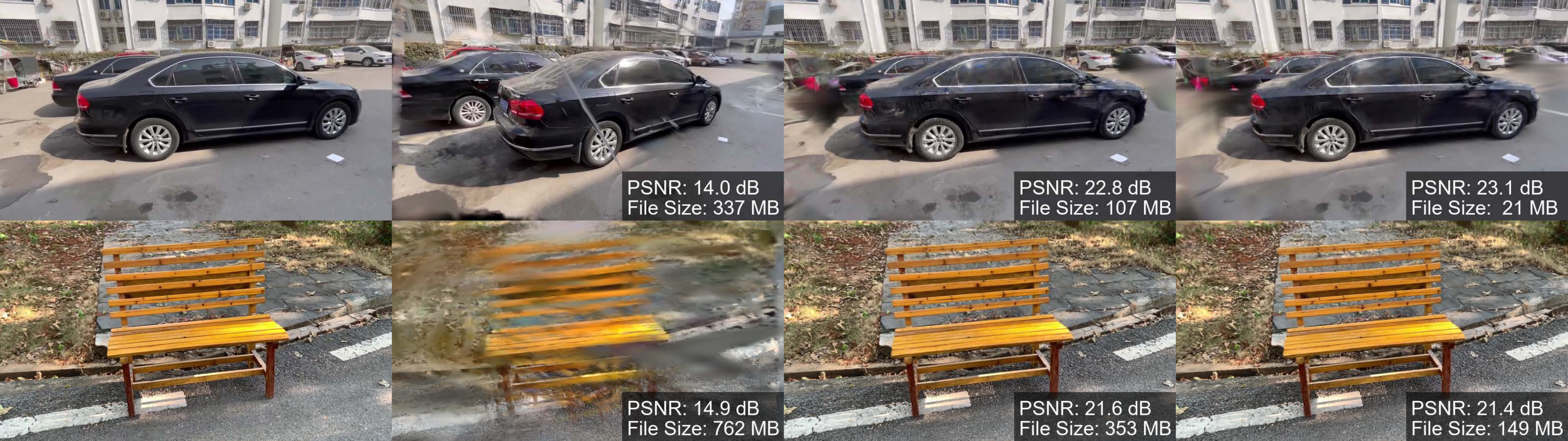} \\

    \raisebox{0.8\height}{\rotatebox{90}{Mip-NeRF 360}} & \rotatebox{90}{\hspace{2em}12 views \hspace{4.5em}3 views}
    \includegraphics[width=0.95\linewidth]{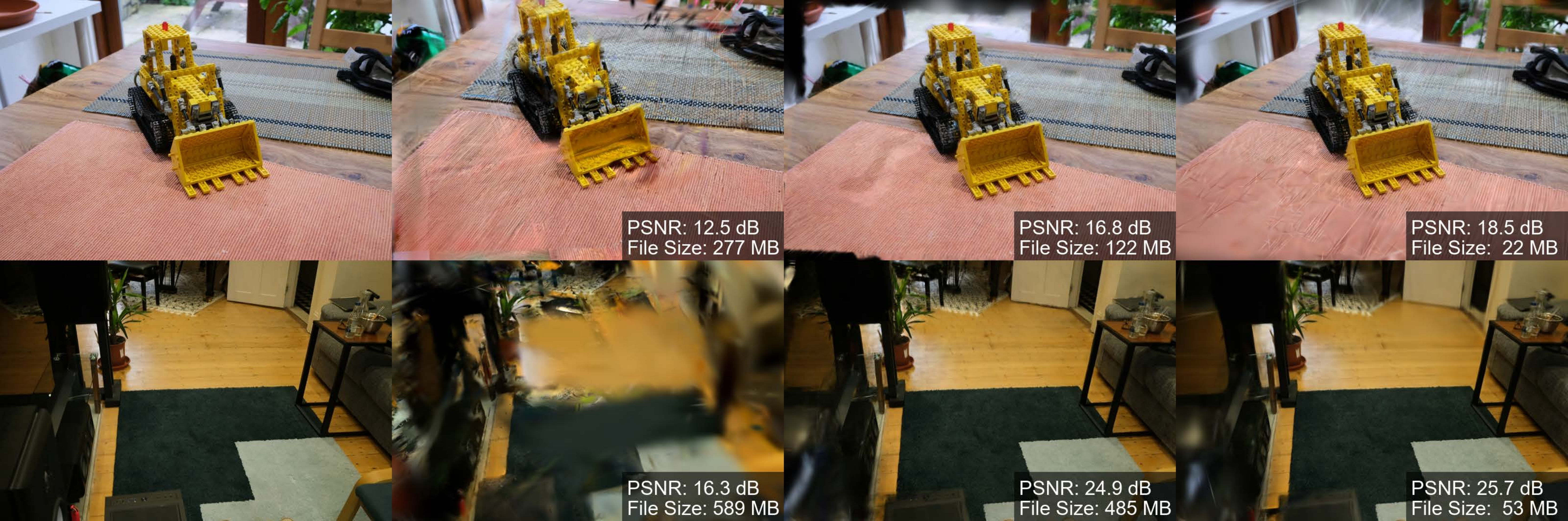} \\
    &
    \begin{minipage}[b]{0.95\linewidth}
        \begin{tabular}{w{c}{40mm}w{c}{39mm}w{c}{39mm}w{c}{39mm}} 
            Ground truth & CF-3DGS \cite{Fu_2024_CVPR} & InstantSplat \cite{fan2024instantsplat} & SDI-GS (Ours)
        \end{tabular}
    \end{minipage}
    
    \end{tabular}
    \caption{Qualitative comparison across three datasets under 3-view (top subrow) and 12-view (bottom subrow) settings. Each row corresponds to a different dataset. Within each dataset, we show rendering results for CF-3DGS, InstantSplat, and our method. CF-3DGS exhibits severe artifacts due to unreliable pose estimation. In contrast, both InstantSplat and our method use MASt3R for initialization and refine poses during training, leading to stable and accurate reconstructions. Our segmentation-driven downsampling further reduces memory usage without compromising visual quality.}
    \label{fig:3}
\end{figure*}

\begin{table*}[t]
\centering
\caption{SfM-Free Comparison on Tanks and Temples across 3, 6, and 12 views}
\label{tab:tnt-horizontal}
\begin{tabular}{ >{\centering\arraybackslash}p{55pt}|
>{\centering\arraybackslash}p{6mm}>{\centering\arraybackslash}p{6mm}>{\centering\arraybackslash}p{6mm}
>{\centering\arraybackslash}p{6mm}>{\centering\arraybackslash}p{6mm}>{\centering\arraybackslash}p{6mm}
>{\centering\arraybackslash}p{6mm}>{\centering\arraybackslash}p{6mm}>{\centering\arraybackslash}p{6mm} 
>{\centering\arraybackslash}p{6mm}>{\centering\arraybackslash}p{6mm}>{\centering\arraybackslash}p{6mm}
>{\centering\arraybackslash}p{6mm}>{\centering\arraybackslash}p{6mm}>{\centering\arraybackslash}p{3mm}}
 \multirow{2}{*}{Method} & \multicolumn{3}{c}{SSIM↑} & \multicolumn{3}{c}{LPIPS↓} & \multicolumn{3}{c}{Size (MB)↓} & \multicolumn{3}{c}{Training Time↓} & \multicolumn{3}{c}{Rendering Speed (FPS)↑}\\
\cline{2-16}
      & 3view& 6view& 12view& 3view& 6view& 12view& 3view& 6view& 12view & 3view & 6view & 12view & 3view & 6view & 12view\\
\hline

CF-3DGS \cite{Fu_2024_CVPR}     & 0.406 & 0.469 & 0.507 & 0.452 & 0.421 & 0.418 & 321.1 & 590.4 & 800.2 & 1m19s & 2m55s & 5m6s  & 156 & 96  & 83\\
InstantSplat \cite{fan2024instantsplat} & \textbf{0.768} & \textbf{0.846} & \textbf{0.866} & \textbf{0.175} & \textbf{0.137} & \textbf{0.139} & 98.6  & 168.1 & 278.4 & 7.12s  & 10.12s & 12.88s & 152 & 131 & 105\\
SDI-GS (Ours)         & 0.754 & 0.831 & 0.853 & 0.245 & 0.192 & 0.178 & \textbf{21.8}  & \textbf{46.3}  & \textbf{85.2}  & \textbf{6.50s}  &  \textbf{8.62s} & \textbf{10.00s} & \textbf{190} & \textbf{173} & \textbf{148}\\

\end{tabular}
\label{tab:1}
\end{table*}

\begin{table*}[t]
\centering
\caption{SfM-free Comparison on Mip-NeRF 360 across 3, 6, and 12 views}
\label{tab:tnt-horizontal}
\begin{tabular}{ >{\centering\arraybackslash}p{55pt}|
>{\centering\arraybackslash}p{6mm}>{\centering\arraybackslash}p{6mm}>{\centering\arraybackslash}p{6mm}
>{\centering\arraybackslash}p{6mm}>{\centering\arraybackslash}p{6mm}>{\centering\arraybackslash}p{6mm}
>{\centering\arraybackslash}p{6mm}>{\centering\arraybackslash}p{6mm}>{\centering\arraybackslash}p{6mm} 
>{\centering\arraybackslash}p{6mm}>{\centering\arraybackslash}p{6mm}>{\centering\arraybackslash}p{6mm}
>{\centering\arraybackslash}p{6mm}>{\centering\arraybackslash}p{6mm}>{\centering\arraybackslash}p{3mm}}
 \multirow{2}{*}{Method} & \multicolumn{3}{c}{SSIM↑} & \multicolumn{3}{c}{LPIPS↓} & \multicolumn{3}{c}{Size (MB)↓} & \multicolumn{3}{c}{Training Time↓} & \multicolumn{3}{c}{Rendering Speed (FPS)↑}\\
\cline{2-16}
      & 3view& 6view& 12view& 3view& 6view& 12view& 3view& 6view& 12view & 3view & 6view & 12view & 3view & 6view & 12view\\
\hline

CF-3DGS \cite{Fu_2024_CVPR}     & 0.217 & 0.256 & 0.244 & 0.594 & 0.599 & 0.609 & 337.27 & 715.09 & 968.58 & 1m46s & 3m57s & 7m7s  & 131 & 87 & 69\\
InstantSplat \cite{fan2024instantsplat} & 0.317 & 0.417 & 0.467 & \textbf{0.534} & \textbf{0.466} & \textbf{0.441} & 118.88 & 227.74 & 429.53 & 16.57s & 19.86s & 25.43s & 115 & 95 & 70\\
SDI-GS (Ours)         & \textbf{0.336} & \textbf{0.431} & \textbf{0.476} & 0.569 & 0.510 & 0.473 & \textbf{20.48}  & \textbf{37.29}  & \textbf{71.99}  & \textbf{16.14s} & \textbf{18.29s} & \textbf{20.86s} & \textbf{138} & \textbf{125} & \textbf{101}\\

\end{tabular}
\label{tab:2}
\end{table*}

\begin{table*}[t]
\centering
\caption{SfM-free Comparison on MVImgNet across 3, 6, and 12 views}
\label{tab:tnt-horizontal}
\begin{tabular}{ >{\centering\arraybackslash}p{55pt}|
>{\centering\arraybackslash}p{6mm}>{\centering\arraybackslash}p{6mm}>{\centering\arraybackslash}p{6mm}
>{\centering\arraybackslash}p{6mm}>{\centering\arraybackslash}p{6mm}>{\centering\arraybackslash}p{6mm}
>{\centering\arraybackslash}p{6mm}>{\centering\arraybackslash}p{6mm}>{\centering\arraybackslash}p{6mm} 
>{\centering\arraybackslash}p{6mm}>{\centering\arraybackslash}p{6mm}>{\centering\arraybackslash}p{6mm}
>{\centering\arraybackslash}p{6mm}>{\centering\arraybackslash}p{6mm}>{\centering\arraybackslash}p{3mm}}
 \multirow{2}{*}{Method} & \multicolumn{3}{c}{SSIM↑} & \multicolumn{3}{c}{LPIPS↓} & \multicolumn{3}{c}{Size (MB)↓} & \multicolumn{3}{c}{Training Time↓} & \multicolumn{3}{c}{Rendering Speed (FPS)↑}\\
\cline{2-16}
      & 3view& 6view& 12view& 3view& 6view& 12view& 3view& 6view& 12view & 3view & 6view & 12view & 3view & 6view & 12view\\
\hline

CF-3DGS \cite{Fu_2024_CVPR}& 0.341 & 0.383 & 0.431 & 0.552 & 0.539 & 0.559 & 413.61 & 766.60 & 708 & 3m16s & 7m24s & 13m5s & 90  & 63  & 70\\
InstantSplat \cite{fan2024instantsplat} & \textbf{0.554} & \textbf{0.694} & \textbf{0.720} & \textbf{0.386} & \textbf{0.283} & \textbf{0.280} & 102.71 & 195.88 & 347 & 8.00s     & 13.33s & 19.56s & 130 & 112 & 76\\
SDI-GS (Ours)         & 0.550 & 0.671 & 0.700 & 0.438 & 0.347 & 0.332 & \textbf{25.53}  & \textbf{55.91}  & \textbf{104} & \textbf{7.00s}     & \textbf{10.78s} & \textbf{14.11s} & \textbf{188} & \textbf{176} & \textbf{145}\\

\end{tabular}
\label{tab:3}
\end{table*}



\begin{table*}[t]
\centering
\caption{SfM-based Comparison on Mip-NeRF 360 and DTU}
\label{tab:sfm}
\begin{tabular}{w{c}{50pt}| w{c}{20pt} w{c}{20pt} w{c}{20pt} w{c}{25pt} w{c}{33pt} w{c}{33pt}| w{c}{20pt} w{c}{20pt} w{c}{20pt} w{c}{25pt} w{c}{33pt} w{c}{33pt}}
\multirow{2}{*}{Method} & \multicolumn{6}{c|}{DTU (3 views)} & \multicolumn{6}{c}{Mip-NeRF 360 (12 views)}\\
\cline{2-13}
& PSNR↑ & SSIM↑ & LPIPS↓ & Size (MB) & Train Time & Render FPS & PSNR↑ & SSIM↑ & LPIPS↓ & Size (MB) & Train Time & Render FPS \\
\hline
FSGS \cite{zhu2023FSGS}      &  20.39 & \textbf{0.827} & 0.206 & 21  & 11m37s & \textbf{141} & 17.49 & 0.582 & \textbf{0.459} & 142 & 13m35s & 28 \\
SparseGS \cite{xiong2023sparsegs}   &  19.89 & 0.778 & \textbf{0.205} & \textbf{16}  & 11m41s & 125 & 16.72 & 0.550 & 0.463 & 134 & 44m30s & 36 \\
CoR-GS  \cite{zhang2024corgs}    &  \textbf{21.98} & \textbf{0.827} & 0.222 & 34  & 7m23s  & 138 & 16.95 & 0.559 & 0.515 & \textbf{40}  & 41m36s & 37 \\
InstantSplat \cite{fan2024instantsplat}&  20.92 & 0.781 & 0.210 & 142 & 34s    & 88  & 17.26 & 0.606 & 0.491 & 465 & 2m23s  & 23 \\
SDI-GS (Ours)        &  21.35 & 0.793 & 0.224 & 24  & \textbf{25s}    & 134 & \textbf{18.50} & \textbf{0.648} & 0.501 & 59  & \textbf{1m16s}  & \textbf{43} \\

\end{tabular}
\label{tab:4}
\end{table*}

\begin{figure}
    \centering
    \includegraphics[width=1.0\linewidth]{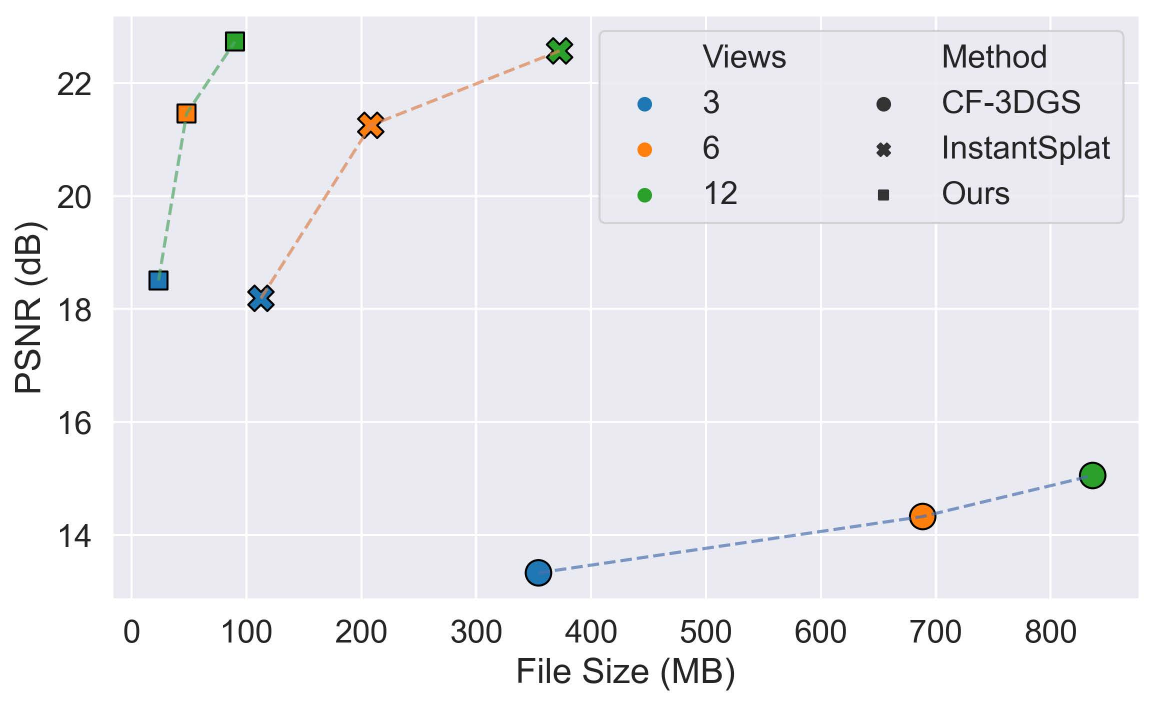}
    \caption{Compression-performance trend across increasing view counts (3, 6, 12), averaged over Tanks and Temples, MVImgNet, and Mip-NeRF 360. Each line shows PSNR versus file size for a method. As the number of input views increases, our method achieves greater compression gains—reducing file size by up to 75\% at 12 views—while maintaining comparable reconstruction quality.}
    \label{fig:4}
    \vspace{-1em}
\end{figure}

\section{Experimental setup}
We evaluated our method (SDI-GS) under SfM-free and SfM-based protocols, using diverse datasets and consistent benchmarks. This enabled fair comparison against baselines under varying sparse-view conditions.

\subsection{Datasets}
\subsubsection*{SfM-free}
We followed the InstantSplat \cite{fan2024instantsplat} protocol, evaluating on eight Tanks and Temples \cite{tankandtample} scenes, seven diverse MVImgNet \cite{yu2023mvimgnet} scenes, and all nine Mip-NeRF 360 scenes. For Tanks and Temples and MVImgNet, 24 images were uniformly sampled per scene. For Mip-NeRF 360, 24 images were drawn from the first 48 frames, capturing objects from varying elevations along a $360^\circ$ trajectory. Training was performed on sparse subsets of $N = {3,6,12}$ views, with the remaining 12 views (excluding first and last frames) reserved for testing, consistent with InstantSplat \cite{fan2024instantsplat}.
\subsubsection*{SfM-based}
The SfM-based evaluation shared the Mip-NeRF 360 dataset but adopts a different training and testing split in line with RegNeRF \cite{Niemeyer2021Regnerf} and RegSegField \cite{kai2024regsegfield} protocols. We selected five Mip-NeRF 360 scenes with clear object boundaries, training on 12 views and testing every 8th image, following community standards. Additionally, we used ten DTU scans with multiple foreground objects, training on 3 views focused on foreground reconstruction.

This distinction in Mip-NeRF 360 splits ensured internal consistency within each protocol and prevents overinterpretation when comparing SfM-free and SfM-based methods. To further explore cross-protocol robustness, we also evaluated InstantSplat under the SfM-based split, highlighting how camera pose quality and view selection influence generalizability.

\subsection{Baselines}
\subsubsection*{SfM-free}
We compared our method against InstantSplat~\cite{instantngp} and CF-3DGS~\cite{Fu_2024_CVPR}, using identical datasets and view selections to ensure a fair comparison.
\subsubsection*{SfM-based}
SfM-based comparisons included FSGS~\cite{zhu2023FSGS}, SparseGS~\cite{xiong2023sparsegs}, and CoR-GS~\cite{zhang2024corgs}. We also included InstantSplat under this protocol to enable direct comparison across SfM paradigms.

\subsection{Implementation Details}
\subsubsection*{SfM-free}
We set training iterations for our method and InstantSplat to 300, aligning with InstantSplat’s protocol, emphasizing rapid convergence from dense initializations. CF-3DGS followed its default two-stage schedule: 300 iterations for local Gaussian optimization, then 300 for global refinement. For evaluation, we used the same resolution as in training to maintain consistency. All SfM-free experiments ran on A100 GPUs.

\subsubsection*{SfM-based}
Training iterations for our method were increased to 1000 on Mip-NeRF 360, reflecting the complexity of outdoor 360° scenes and the slower convergence typical of SfM-based methods. On the simpler DTU dataset, we kept 300 iterations. Baselines FSGS, SparseGS, and CoR-GS used their default settings: 30,000 iterations on Mip-NeRF 360, and on DTU, 30,000 for FSGS and SparseGS, and 10,000 for CoR-GS.

Evaluation generally used each method’s training resolution, except for Mip-NeRF 360 SfM-based runs, where we tested on the original high-resolution images rather than downsampled inputs. This provided a more challenging benchmark for high-fidelity view synthesis, albeit with slower rendering speeds compared to DTU and SfM-free settings. All SfM-based experiments used A40 GPUs.

\begin{figure*}
    \centering
    \begin{tabular}{cc}
    \raisebox{0.4\height}{\rotatebox{90}{Mip-NeRF 360}} &
    \includegraphics[width=0.95\linewidth]{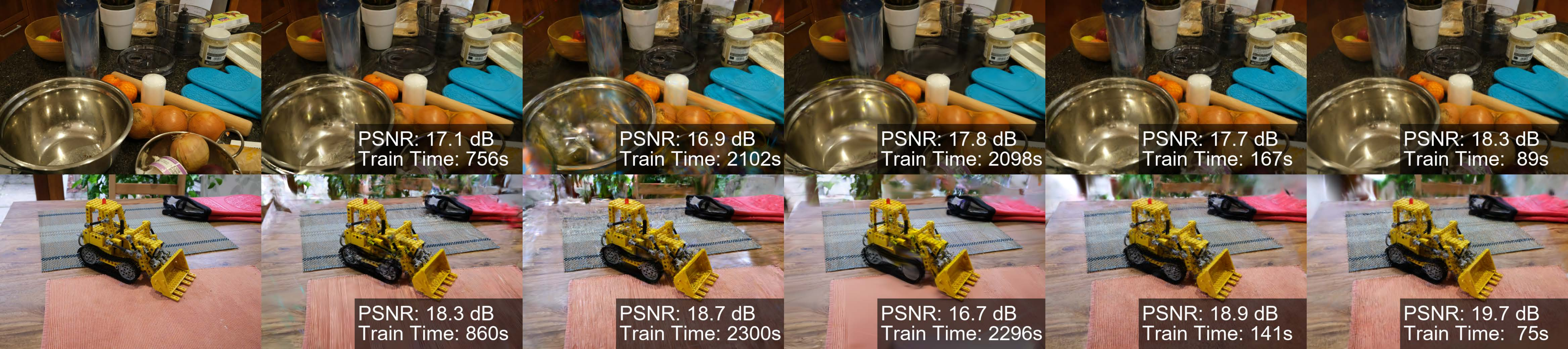} \\
    
    \raisebox{2.5\height}{\rotatebox{90}{DTU}} &
    \includegraphics[width=0.95\linewidth]{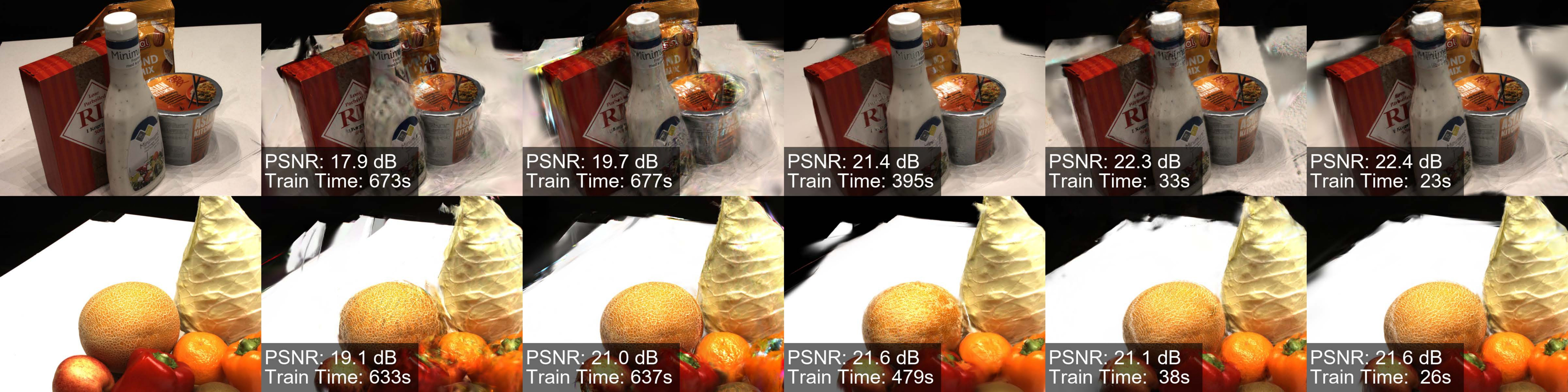} \\
    
    &
    \begin{minipage}[b]{0.95\linewidth}
        \begin{tabular}{w{c}{25mm}w{c}{24mm}w{c}{24mm}w{c}{25mm}w{c}{24mm}w{c}{24mm}} 
            Ground truth & FSGS \cite{zhu2023FSGS} & SparseGS \cite{xiong2023sparsegs}& CoR-GS \cite{zhang2024corgs}& InstantSplat \cite{fan2024instantsplat}& SDI-GS (Ours)
        \end{tabular}
    \end{minipage}
    
    \end{tabular}
    \caption{Qualitative comparison with SfM-based methods (FSGS, SparseGS, CoR-GS) and SfM-free InstantSplat. SDI-GS matches the visual quality of SfM-based pipelines while keeping the lightweight memory footprint of SfM-free methods (Table \ref{tab:4}) and drastically reducing training time. Unlike InstantSplat, which demands more storage, our segmentation-driven initialization provides a compact, efficient representation without compromising structural fidelity.}
    \label{fig:5}
\end{figure*}


\section{Results}

We report results under the SfM-free and SfM-based settings described in Section 4. The following subsections present detailed comparisons with state-of-the-art methods in each category.

\subsection{SfM-free Comparison}



Tables~\ref{tab:1}, \ref{tab:2}, and \ref{tab:3} present quantitative results on Tanks and Temples, Mip-NeRF 360, and MVImgNet under varying sparse-view settings (3, 6, and 12 views), respectively. Across all datasets, our method achieves comparable SSIM and only marginally worse LPIPS scores compared to InstantSplat, while significantly reducing the memory footprint. On Tanks and Temples, our file size is reduced to one-third of InstantSplat’s, and we match its rendering quality with faster training and 25\% lower rendering latency. On Mip-NeRF 360, our method attains slightly better SSIM while LPIPS remains competitive; notably, we reduce file size by over 80\%. On MVImgNet, our rendering quality is slightly lower, but we maintain a 3× smaller model and up to 40\% faster rendering speed.

Visual comparisons are shown in Fig.~\ref{fig:3}, which display 3-view and 12-view rendering results. In the 3-view case, CF-3DGS exhibits severe artifacts due to inaccurate camera pose estimation, often leading to completely misaligned views. Notably, its progressive densification strategy does not guarantee improved quality with more views; additional training views may propagate incorrect geometry due to faulty pose initialization, worsening results in some cases. In contrast, InstantSplat and our method both use MASt3R for pose initialization and jointly optimize poses during training, resulting in more accurate and stable reconstructions.

Our results preserve global structure well compared to InstantSplat, though we may miss some high-frequency details. Importantly, our segmentation-guided downsampling smooths the transitions in overlapping regions—areas where InstantSplat tends to exhibit abrupt visual artifacts. Moreover, our method achieves these results with significantly lower memory usage: requiring only one-third the memory of InstantSplat in 3-view setups and up to 80\% less in the 12-view configuration (Table \ref{tab:3}). 

Fig.~\ref{fig:4} provides a deeper analysis of how compression efficiency scales with the number of input views. As the number of training views increases from 3 to 12, so does the redundancy in the lifted point clouds. Our segmentation-guided downsampling capitalizes on this redundancy, yielding more aggressive compression. Specifically, we observe a 50\% reduction in file size for 3-view inputs, and up to 75\% for 12 views compared to InstantSplat. Importantly, this is achieved without compromising reconstruction quality, as shown by the maintained PSNR. This trend underscores the scalability of our method: the more complex the input, the greater the efficiency gain, all while preserving structural fidelity in the final representation.

These results demonstrate the practical effectiveness of our method in SfM-free scenarios, especially for compact and real-time applications in sparse-view settings. These results demonstrate the practical efficiency and effectiveness of our method for SfM-free sparse-view synthesis. Our method enables high-quality rendering with significantly lower memory and time costs, which is ideal for real-time and resource-constrained scenarios.

\subsection{SfM-based Comparison}

Table \ref{tab:4} summarizes the comparison of our method with SfM-based baselines. These baselines begin with sparse point clouds generated by SfM and therefore rely on explicit densification to reach sufficient coverage. This procedure is not required in SfM-free methods, which begin with dense pixel-wise reconstructions. As a result, SfM-based pipelines typically require significantly more training—often 10× longer—to progressively grow the Gaussian representation. In contrast, our method achieves fast convergence by directly training on filtered dense point clouds.

Our method achieves comparable quantitative performance to SfM-based baselines on the DTU dataset, and outperforms them in both PSNR and SSIM on Mip-NeRF 360, albeit with a slightly higher LPIPS. In terms of file size, we match or exceed the compactness of SfM-based methods. On DTU, these methods benefit from the sparsity of the foreground-centric scenes, resulting in lower file sizes. However, InstantSplat incurs a large memory overhead due to projecting all pixels—including background—into 3D, even when such regions are low in structural relevance. Our segmentation-driven downsampling addresses this redundancy, resulting in a much more compact representation while preserving fidelity.

We now compare efficiency across methods in terms of memory footprints and rendering speed. On Mip-NeRF 360, SfM-based methods require longer training to achieve comparable reconstruction quality, which leads to larger Gaussian counts and slower rendering. Only CoR-GS applies a Gaussian pruning strategy to reduce the final memory size. However, its dual-Gaussian design doubles memory consumption during training and maintains a high computational cost. In contrast, our method benefits from initializing with downsampled dense predictions, reducing file size to nearly half that of FSGS and SparseGS, and to one-eighth that of InstantSplat. Our rendering speed is faster than all baselines on Mip-NeRF 360, and competitive with FSGS on DTU.

Fig.~\ref{fig:5} presents qualitative comparisons on representative scenes. While maintaining visual quality comparable to SfM-based methods, our approach offers drastically faster training by orders of magnitude and a more compact model. 

These results underscore that our method bridges the gap between SfM-free and SfM-based pipelines. By leveraging dense initialization while enforcing compactness through structured downsampling, we combine the efficiency of SfM-free training with the lightweight output of SfM-based pipelines. Our method achieves low memory and time cost without sacrificing rendering quality.

\subsection{Ablation study}

Our method uses a 3D label vector constructed from region-based segmentations across three input views. This design balances computational efficiency with segmentation consistency. To validate this design choice, we conduct an ablation study comparing against a variant that uses various available labels from the 12-view setup (i.e., a label dimension of 6 and 12). Since higher label dimensions yield more segments and therefore more Gaussians, we increase training to 1000 iterations to ensure convergence across all settings.

Table~\ref{tab:5} shows that increasing the label dimension from 3 to 12 leads to a significant growth in file size—with only marginal improvements in PSNR and SSIM. This is because higher label dimensionality introduces many fragmented or noisy segment combinations, breaking coherent structures into smaller, less meaningful clusters. These fine-grained segments inflate the number of retained Gaussians, undermining the intended sparsity without substantial benefit to reconstruction quality. 


\begin{table}[t]
\centering
\caption{Ablation study on label dimension under the 12-view setting.}
\begin{tabular}{w{c}{30pt} w{c}{15pt} w{c}{15pt} w{c}{18pt} w{c}{30pt} w{c}{30pt} w{c}{30pt}}

Dimension & PSNR↑ & SSIM↑ & LPIPS↓ & Size (MB) & Train Time & Render FPS \\
\hline
3   & \textbf{17.83} & \textbf{0.445} & 0.452 & \textbf{72}  & \textbf{44s} & \textbf{161} \\
6   & 17.42 & 0.434 & 0.442 & 154 & 55s & 116 \\
12  & 17.28 & 0.419 & \textbf{0.438} & 247 & 67s & 89 \\
\end{tabular}
\label{tab:5}
\end{table}

\section{Conclusion}
We have presented a segmentation-driven initialization strategy for 3D Gaussian Splatting in sparse-view settings, addressing the inefficiencies present in existing SfM-free methods. Rather than lifting every pixel into 3D space indiscriminately, our method applies region-based segmentation to guide structured downsampling, significantly reducing redundancy while preserving scene fidelity.

Through extensive experiments on diverse datasets, we have demonstrated that our SDI-GS achieves comparable or superior rendering quality to existing SfM-free and SfM-based methods, with up to 50\% fewer Gaussians and substantially lower memory and runtime cost. Notably, SDI-GS preserves the fast training times that make SfM-free pipelines attractive, while producing representations as compact as those of SfM-based methods.

These results highlight the importance of structure-aware initialization in 3DGS. By reducing unnecessary overhead at the source, we improve scalability and practicality under real-world sparse-view constraints. Future directions include exploring adaptive or learned segmentation techniques and extending our framework to unconstrained, dynamic, or low-light environments.

\bibliographystyle{IEEEtran}
\bibliography{refs}

@inproceedings{sarlin20superglue,
  author    = {Paul-Edouard Sarlin and
               Daniel DeTone and
               Tomasz Malisiewicz and
               Andrew Rabinovich},
  title     = {{SuperGlue}: Learning Feature Matching with Graph Neural Networks},
  booktitle = {CVPR},
  year      = {2020},
  url       = {https://arxiv.org/abs/1911.11763}
}

@inproceedings{mapnet2018,
  title={Geometry-Aware Learning of Maps for Camera Localization},
  author={Samarth Brahmbhatt and Jinwei Gu and Kihwan Kim and James Hays and Jan Kautz},
  booktitle={IEEE Conference on Computer Vision and Pattern Recognition (CVPR)},
  year={2018}
}

@INPROCEEDINGS {kendall2015posenet,
author = { Kendall, Alex and Grimes, Matthew and Cipolla, Roberto },
booktitle = { 2015 IEEE International Conference on Computer Vision (ICCV) },
title = {{ PoseNet: A Convolutional Network for Real-Time 6-DOF Camera Relocalization }},
year = {2015},
volume = {},
ISSN = {2380-7504},
pages = {2938-2946},
doi = {10.1109/ICCV.2015.336},
url = {https://doi.ieeecomputersociety.org/10.1109/ICCV.2015.336},
publisher = {IEEE Computer Society},
address = {Los Alamitos, CA, USA},
month =Dec}

@inproceedings{mast3r,
author = {Leroy, Vincent and Cabon, Yohann and Revaud, Jerome},
title = {Grounding Image Matching in 3D with MASt3R},
year = {2024},
isbn = {978-3-031-73219-5},
publisher = {Springer-Verlag},
address = {Berlin, Heidelberg},
doi = {10.1007/978-3-031-73220-1_5},
booktitle = {Computer Vision – ECCV 2024: 18th European Conference, Milan, Italy, September 29–October 4, 2024, Proceedings, Part LXXII},
pages = {71–91},
numpages = {21},
location = {Milan, Italy}
}

@misc{zhu20243dgsrobotics,
      title={3D Gaussian Splatting in Robotics: A Survey}, 
      author={Siting Zhu and Guangming Wang and Xin Kong and Dezhi Kong and Hesheng Wang},
      year={2024},
      eprint={2410.12262},
      archivePrefix={arXiv},
      primaryClass={cs.RO},
      url={https://arxiv.org/abs/2410.12262}, 
}

@article{zhang2023sam3d,
  title={SAM3D: Zero-Shot 3D Object Detection via Segment Anything Model},
  author={Zhang, Dingyuan and Liang, Dingkang and Yang, Hongcheng and Zou, Zhikang and Ye, Xiaoqing and Liu, Zhe and Bai, Xiang},
  journal={Science China Information Sciences},
  year={2023}
}

@inproceedings{ye2024gaussian_grouping,
    title={Gaussian Grouping: Segment and Edit Anything in 3D Scenes},
    author={Ye, Mingqiao and Danelljan, Martin and Yu, Fisher and Ke, Lei},
    booktitle={ECCV},
    year={2024}
}

@article{adptsmoe,
  title={Adaptive Segmentation-Based Initialization for Steered-Mixture-of-Experts Image Regression},
  author={Li, Yi-Hsin and Knorr, Sebastian and Sjöström, Mårten and Sikora, Thomas},
  journal={IEEE Transaction on Multimedia},
   year={2025},
}

@inproceedings{li_segmentation-based_2023,
address = {Jeju, Korea (South)},
	author={Li, Yi-Hsin and Sjöström, Mårten and Knorr, Sebastian and Sikora, Thomas},
  booktitle={2023 IEEE International Conference on Visual Communications and Image Processing (VCIP)}, 
  title={Segmentation-based Initialization for Steered Mixture of Experts}, 
  year={2023},
  volume={},
  number={},
  pages={1-5},
  doi={10.1109/VCIP59821.2023.10402643}}

@InProceedings{zhu2024endogs,
author="Zhu, Lingting and Wang, Zhao and Cui, Jiahao and Jin, Zhenchao and Lin, Guying and Yu, Lequan",
editor="Celebi, M. Emre
and Reyes, Mauricio
and Chen, Zhen
and Li, Xiaoxiao",
title="EndoGS: Deformable Endoscopic Tissues Reconstruction with Gaussian Splatting",
booktitle="Medical Image Computing and Computer Assisted Intervention -- MICCAI 2024 Workshops",
year="2025",
publisher="Springer Nature Switzerland",
address="Cham",
pages="135--145",
isbn="978-3-031-77610-6"
}

@misc{zhou2023tiavox,
      title={TiAVox: Time-aware Attenuation Voxels for Sparse-view 4D DSA Reconstruction}, 
      author={Zhenghong Zhou and Huangxuan Zhao and Jiemin Fang and Dongqiao Xiang and Lei Chen and Lingxia Wu and Feihong Wu and Wenyu Liu and Chuansheng Zheng and Xinggang Wang},
      year={2023},
      eprint={2309.02318},
      archivePrefix={arXiv},
      primaryClass={cs.CV},
      url={https://arxiv.org/abs/2309.02318}, 
}

@inproceedings{jiang2024vrgs,
author = {Jiang, Ying and Yu, Chang and Xie, Tianyi and Li, Xuan and Feng, Yutao and Wang, Huamin and Li, Minchen and Lau, Henry and Gao, Feng and Yang, Yin and Jiang, Chenfanfu},
title = {VR-GS: A Physical Dynamics-Aware Interactive Gaussian Splatting System in Virtual Reality},
year = {2024},
isbn = {9798400705250},
publisher = {Association for Computing Machinery},
address = {New York, NY, USA},
url = {https://doi.org/10.1145/3641519.3657448},
doi = {10.1145/3641519.3657448},
booktitle = {ACM SIGGRAPH 2024 Conference Papers},
articleno = {78},
numpages = {1},
location = {Denver, CO, USA},
series = {SIGGRAPH '24}
}

@INPROCEEDINGS {meuleman2023progressivenerf,
author = { Meuleman, Andreas and Liu, Yu-Lun and Gao, Chen and Huang, Jia-Bin and Kim, Changil and Kim, Min H. and Kopf, Johannes },
booktitle = { 2023 IEEE/CVF Conference on Computer Vision and Pattern Recognition (CVPR) },
title = {Progressively Optimized Local Radiance Fields for Robust View Synthesis},
year = {2023},
volume = {},
ISSN = {},
pages = {16539-16548},
doi = {10.1109/CVPR52729.2023.01587},
url = {https://doi.ieeecomputersociety.org/10.1109/CVPR52729.2023.01587},
publisher = {IEEE Computer Society},
address = {Los Alamitos, CA, USA},
month =Jun}

@InProceedings{Kirillov2023SAM,
    author    = {Kirillov, Alexander and Mintun, Eric and Ravi, Nikhila and Mao, Hanzi and Rolland, Chloe and Gustafson, Laura and Xiao, Tete and Whitehead, Spencer and Berg, Alexander C. and Lo, Wan-Yen and Dollar, Piotr and Girshick, Ross},
    title     = {Segment Anything},
    booktitle = {Proceedings of the IEEE/CVF International Conference on Computer Vision (ICCV)},
    month     = {October},
    year      = {2023},
    pages     = {4015-4026}
}

@misc{zhu2023FSGS,
    title={FSGS: Real-Time Few-Shot View Synthesis using Gaussian Splatting},
    author={Zehao Zhu and Zhiwen Fan and Yifan Jiang and Zhangyang Wang},
    year={2023},
    eprint={2312.00451},
    archivePrefix={arXiv},
    primaryClass={cs.CV}
}

@article{xiong2023sparsegs,
  author    = {Xiong, Haolin and Muttukuru, Sairisheek and Upadhyay, Rishi and Chari, Pradyumna and Kadambi, Achuta},
  title     = {SparseGS: Real-Time 360° Sparse View Synthesis using Gaussian Splatting},
  journal   = {Arxiv},
  year      = {2023},
}

@inproceedings{kai2024regsegfield,
author = {Gu, Kai and Maugey, Thomas and Sebastian, Knorr and Guillemot, Christine},
title = {RegSegField: Mask-Regularization and Hierarchical Segmentation for Novel View Synthesis from Sparse Inputs},
year = {2024},
isbn = {9798400712814},
publisher = {Association for Computing Machinery},
address = {New York, NY, USA},
url = {https://doi.org/10.1145/3697294.3697299},
doi = {10.1145/3697294.3697299},
booktitle = {Proceedings of 21st ACM SIGGRAPH Conference on Visual Media Production},
articleno = {3},
numpages = {10},
location = {London, United Kingdom},
series = {CVMP '24}
}

@article{westoby2012sfm,
title = {‘Structure-from-Motion’ photogrammetry: A low-cost, effective tool for geoscience applications},
journal = {Geomorphology},
volume = {179},
pages = {300-314},
year = {2012},
issn = {0169-555X},
doi = {https://doi.org/10.1016/j.geomorph.2012.08.021},
url = {https://www.sciencedirect.com/science/article/pii/S0169555X12004217},
author = {M.J. Westoby and J. Brasington and N.F. Glasser and M.J. Hambrey and J.M. Reynolds}
}

@misc{fan2024instantsplat,
        title={InstantSplat: Sparse-view Gaussian Splatting in Seconds}, 
        author={Zhiwen Fan and Kairun Wen and Wenyan Cong and Kevin Wang and Jian Zhang and Xinghao Ding and Danfei Xu and Boris Ivanovic and Marco Pavone and Georgios Pavlakos and Zhangyang Wang and Yue Wang},
        year={2024},
        eprint={2403.20309},
        archivePrefix={arXiv},
        primaryClass={cs.CV}
      }

@inproceedings{dust3r_cvpr24,
      title={DUSt3R: Geometric 3D Vision Made Easy}, 
      author={Shuzhe Wang and Vincent Leroy and Yohann Cabon and Boris Chidlovskii and Jerome Revaud},
      booktitle = {CVPR},
      year = {2024}
}

@inproceedings{schoenberger2016sfm,
    author={Sch\"{o}nberger, Johannes Lutz and Frahm, Jan-Michael},
    title={Structure-from-Motion Revisited},
    booktitle={Conference on Computer Vision and Pattern Recognition (CVPR)},
    year={2016},
}

@inproceedings{schoenberger2016mvs,
    author={Sch\"{o}nberger, Johannes Lutz and Zheng, Enliang and Pollefeys, Marc and Frahm, Jan-Michael},
    title={Pixelwise View Selection for Unstructured Multi-View Stereo},
    booktitle={European Conference on Computer Vision (ECCV)},
    year={2016},
}

@InProceedings{Fu_2024_CVPR,
                    author    = {Fu, Yang and Liu, Sifei and Kulkarni, Amey and Kautz, Jan and Efros, Alexei A. and Wang, Xiaolong},
                    title     = {COLMAP-Free 3D Gaussian Splatting},
                    booktitle = {Proceedings of the IEEE/CVF Conference on Computer Vision and Pattern Recognition (CVPR)},
                    month     = {June},
                    year      = {2024},
                    pages     = {20796-20805}
                }

@InProceedings{truong2023SPARF,
    author    = {Truong, Prune and Rakotosaona, Marie-Julie and Manhardt, Fabian and Tombari, Federico},
    title     = {SPARF: Neural Radiance Fields From Sparse and Noisy Poses},
    booktitle = {Proceedings of the IEEE/CVF Conference on Computer Vision and Pattern Recognition (CVPR)},
    month     = {June},
    year      = {2023},
    pages     = {4190-4200}
}

@InProceedings{Niemeyer2021Regnerf,
          author    = {Michael Niemeyer and Jonathan T. Barron and Ben Mildenhall and Mehdi S. M. Sajjadi and Andreas Geiger and Noha Radwan},  
          title     = {RegNeRF: Regularizing Neural Radiance Fields for View Synthesis from Sparse Inputs},
          booktitle = {Proc. IEEE Conf. on Computer Vision and Pattern Recognition (CVPR)},
          year      = {2022},
        }

@InProceedings{chibane2021SRF,
    author    = {Chibane, Julian and Bansal, Aayush and Lazova, Verica and Pons-Moll, Gerard},
    title     = {Stereo Radiance Fields (SRF): Learning View Synthesis for Sparse Views of Novel Scenes},
    booktitle = {Proceedings of the IEEE/CVF Conference on Computer Vision and Pattern Recognition (CVPR)},
    month     = {June},
    year      = {2021},
    pages     = {7911-7920}
}

@inproceedings{zhang2024corgs,
author = {Zhang, Jiawei and Li, Jiahe and Yu, Xiaohan and Huang, Lei and Gu, Lin and Zheng, Jin and Bai, Xiao},
title = {CoR-GS: Sparse-View 3D Gaussian Splatting via Co-regularization},
year = {2024},
isbn = {978-3-031-73231-7},
publisher = {Springer-Verlag},
address = {Berlin, Heidelberg},
doi = {10.1007/978-3-031-73232-4_19},
booktitle = {Computer Vision – ECCV 2024: 18th European Conference, Milan, Italy, September 29–October 4, 2024, Proceedings, Part I},
pages = {335–352},
numpages = {18},
location = {Milan, Italy}
}

@inproceedings{li2024dngaussian,
author = { Li, Jiahe and Zhang, Jiawei and Bai, Xiao and Zheng, Jin and Ning, Xin and Zhou, Jun and Gu, Lin },
booktitle = { 2024 IEEE/CVF Conference on Computer Vision and Pattern Recognition (CVPR) },
title = {{ DNGaussian: Optimizing Sparse-View 3D Gaussian Radiance Fields with Global-Local Depth Normalization }},
year = {2024},
volume = {},
ISSN = {},
pages = {20775-20785},
doi = {10.1109/CVPR52733.2024.01963},
publisher = {IEEE Computer Society},
address = {Los Alamitos, CA, USA},
month =Jun}

@inproceedings{yu2023mvimgnet,
    title     = {MVImgNet: A Large-scale Dataset of Multi-view Images},
    author    = {Yu, Xianggang and Xu, Mutian and Zhang, Yidan and Liu, Haolin and Ye, Chongjie and Wu, Yushuang and Yan, Zizheng and Liang, Tianyou and Chen, Guanying and Cui, Shuguang and Han, Xiaoguang},
    booktitle = {CVPR},
    year      = {2023}
}

@ARTICLE{shen2016dbscan,
  author={Shen, Jianbing and Hao, Xiaopeng and Liang, Zhiyuan and Liu, Yu and Wang, Wenguan and Shao, Ling},
  journal={IEEE Transactions on Image Processing}, 
  title={Real-Time Superpixel Segmentation by DBSCAN Clustering Algorithm}, 
  year={2016},
  volume={25},
  number={12},
  pages={5933-5942},
  doi={10.1109/TIP.2016.2616302}}

@inproceedings{neubert_compact_2014,
	title = {Compact {Watershed} and {Preemptive} {SLIC}: {On} {Improving} {Trade}-offs of {Superpixel} {Segmentation} {Algorithms}},
	shorttitle = {Compact {Watershed} and {Preemptive} {SLIC}},
	doi = {10.1109/ICPR.2014.181},
	booktitle = {2014 22nd {International} {Conference} on {Pattern} {Recognition}},
	author = {Neubert, Peer and Protzel, Peter},
	month = aug,
	year = {2014},
	note = {ISSN: 1051-4651},
	pages = {996--1001},
}

@inproceedings{li_superpixel_2015,
	title = {Superpixel segmentation using {Linear} {Spectral} {Clustering}},
	doi = {10.1109/CVPR.2015.7298741},
	booktitle = {2015 {IEEE} {Conference} on {Computer} {Vision} and {Pattern} {Recognition} ({CVPR})},
	author = {Li, Zhengqin and Chen, Jiansheng},
	month = jun,
	year = {2015},
	pages = {1356--1363}
}

@inproceedings{yang_superpixel_2020,
    address = {Seattle, WA, USA},
	title = {Superpixel {Segmentation} {With} {Fully} {Convolutional} {Networks}},
	isbn = {978-1-72817-168-5},
	url = {https://ieeexplore.ieee.org/document/9156320/},
	doi = {10.1109/CVPR42600.2020.01398},
	language = {en},
	urldate = {2023-07-12},
	booktitle = {2020 {IEEE}/{CVF} {Conference} on {Computer} {Vision} and {Pattern} {Recognition} ({CVPR})},
	publisher = {IEEE},
	author = {Yang, Fengting and Sun, Qian and Jin, Hailin and Zhou, Zihan},
	month = jun,
	year = {2020},
	pages = {13961--13970},
}

@article{zhang_fast_2023,
	title = {Fast and accurate superpixel segmentation algorithm with a guidance image},
	volume = {129},
	issn = {0262-8856},
	doi = {10.1016/j.imavis.2022.104596},
	language = {en},
	urldate = {2023-07-12},
	journal = {Image and Vision Computing},
	author = {Zhang, Yongsheng and Zhang, Yongxia and Fan, Linwei and Wang, Nannan},
	month = jan,
	year = {2023},
	pages = {104596},
}

@article{kerbl_3d_2023,
	title = {{3D} {Gaussian} {Splatting} for {real}-{time} {radiance} {field} {rendering}},
	volume = {42},
	issn = {0730-0301},
	doi = {10.1145/3592433},
	number = {4},
	urldate = {2024-03-27},
	journal = {ACM Transactions on Graphics},
	author = {Kerbl, Bernhard and Kopanas, Georgios and Leimkuehler, Thomas and Drettakis, George},
	month = jul,
	year = {2023},
	pages = {139:1--139:14},
}

@String{Computing = "Computing" }

@String{Computer = "{IEEE} Computer" }

@String{Springer = "Springer-Verlag" }

@article{nerf,
author = {Mildenhall, Ben and Srinivasan, Pratul P. and Tancik, Matthew and Barron, Jonathan T. and Ramamoorthi, Ravi and Ng, Ren},
title = {NeRF: representing scenes as neural radiance fields for view synthesis},
year = {2021},
issue_date = {January 2022},
publisher = {Association for Computing Machinery},
address = {New York, NY, USA},
volume = {65},
number = {1},
issn = {0001-0782},
url = {https://doi.org/10.1145/3503250},
doi = {10.1145/3503250},
journal = {Commun. ACM},
month = dec,
pages = {99–106},
numpages = {8}
}

@article{instantngp,
author = {M\"{u}ller, Thomas and Evans, Alex and Schied, Christoph and Keller, Alexander},
title = {Instant neural graphics primitives with a multiresolution hash encoding},
year = {2022},
issue_date = {July 2022},
publisher = {Association for Computing Machinery},
address = {New York, NY, USA},
volume = {41},
number = {4},
issn = {0730-0301},
url = {https://doi.org/10.1145/3528223.3530127},
doi = {10.1145/3528223.3530127},
journal = {ACM Trans. Graph.},
month = jul,
articleno = {102},
numpages = {15}
}

@article{tankandtample,
author = {Knapitsch, Arno and Park, Jaesik and Zhou, Qian-Yi and Koltun, Vladlen},
title = {Tanks and temples: benchmarking large-scale scene reconstruction},
year = {2017},
issue_date = {August 2017},
publisher = {Association for Computing Machinery},
address = {New York, NY, USA},
volume = {36},
number = {4},
issn = {0730-0301},
url = {https://doi.org/10.1145/3072959.3073599},
doi = {10.1145/3072959.3073599},
journal = {ACM Trans. Graph.},
month = jul,
articleno = {78},
numpages = {13}
}

\vspace{-33pt}
\begin{IEEEbiographynophoto}{Yi-Hsin Li}
She started a double degree PhD program at Technical University Berlin, Germany and seconded to MIUN in November 2021. She in her four-year PhD journey works on high-dimensional data compression, focusing on gating networks. 
\end{IEEEbiographynophoto}
\vspace{-33pt}
\begin{IEEEbiographynophoto}{Sebastian~Knorr}
Sebastian Knorr, professor of Visual Computing at HTW Berlin. Broad competence in 3D/ 360° imaging technologies. Expertise in 1) capture and synthesis of stereo 3D and 360° video; 2) visual attention and quality assessment in VR; 3) light field imaging and neural radiance fields. 
\end{IEEEbiographynophoto}
\vspace{-33pt}
\begin{IEEEbiographynophoto}{Mårten~Sjöström}
Mårten Sjöström, professor of signal processing, Mid Sweden University. Multidimensional signal processing, imaging and compression; system modelling and identification; inverse problems using machine learning. 120+ scientific articles, two book chapters. Supervisor of 8 PhD students, 8 previously.
\end{IEEEbiographynophoto}
\vspace{-33pt}
\begin{IEEEbiographynophoto}{Thomas~Sikora}
Thomas Sikora, professor and director of the Communication Systems Lab, Technische Universität Berlin, Germany.
\end{IEEEbiographynophoto}

\end{document}